\documentclass[11pt,a4paper]{article}
\usepackage[hyperref]{ranlp2023}
\usepackage{times}
\usepackage{latexsym}

\usepackage{microtype}

\usepackage{booktabs}
\usepackage[T1]{fontenc}

\usepackage[utf8]{inputenc}

\usepackage{microtype}

\usepackage{inconsolata}

\usepackage{graphicx}
\usepackage{dirtytalk}
\usepackage{multirow}
\usepackage{enumitem}

\aclfinalcopy 

\title{Seeing, Signing, and Saying: A Vision-Language Model-Assisted Pipeline for Sign Language Data Acquisition and Curation from Social Media}

\author{
Shakib Yazdani$^{1}$ 
Yasser Hamidullah$^{1}$ 
Cristina España-Bonet$^{1,2}$ 
Josef van Genabith$^{1}$ \\
\texttt{\{shakib.yazdani,yasser.hamidullah,cristinae,josef.van\_genabith\}@dfki.de} \\
$^{1}$German Research Center for Artificial Intelligence (DFKI GmbH),\\
Saarland Informatics Campus, Saarbrücken, Germany \\
$^{2}$Barcelona Supercomputing Center (BSC-CNS), Barcelona, Catalonia, Spain
}

\date{}

\begin{document}
\maketitle
\begin{abstract}

Most existing sign language translation (SLT) datasets are limited in scale, lack multilingual coverage, and are costly to curate due to their reliance on expert annotation and controlled recording setup. Recently, Vision Language Models (VLMs) have demonstrated strong capabilities as evaluators and real-time assistants. Despite these advancements, their potential remains untapped in the context of sign language dataset acquisition. To bridge this gap, we introduce the first automated annotation and filtering framework that utilizes VLMs to reduce reliance on manual effort while preserving data quality. Our method is applied to TikTok videos across eight sign languages and to the already curated YouTube-SL-25 dataset in German Sign Language for the purpose of additional evaluation. Our VLM-based pipeline includes a face visibility detection, a sign activity recognition, a text extraction from video content, and a judgment step to validate alignment between video and text, implementing generic filtering, annotation and validation steps. Using the resulting corpus, TikTok-SL-8, we assess the performance of two off-the-shelf SLT models on our filtered dataset for German and American Sign Languages, with the goal of establishing baselines and evaluating the robustness of recent models on automatically extracted, slightly noisy data. Our work enables scalable, weakly supervised pretraining for SLT and facilitates data acquisition from social media.\footnote{The dataset (TikTok video IDs) and code for our pipeline is available at \url{https://github.com/shakibyzn/vlm-sign-curator}.}

\end{abstract}

\section{Introduction}


Sign languages are rich, expressive visual languages that serve as the primary means of communication for millions of Deaf and Hard-of-Hearing individuals worldwide. Sign languages convey meaning through a combination of hand configurations, movements, facial expressions, and body posture. Each sign language follows its own unique grammar and structure, making them distinct from one another \citep{stokoe1980sign}. 

Sign language translation (SLT), the task of translating a sign language video into spoken language, is evolving both in terms of dataset scale and the performance of SLT models. Despite the increasing availability of larger sign language datasets, such as YouTube-ASL \citep{10.5555/3666122.3667386} for American Sign Language (ASL) and BOBSL \citep{Albanie2021bobsl} for British Sign Language (BSL), as well as multilingual resources like JWSign \citep{gueuwou-etal-2023-jwsign} or SP-10 \citep{yin_sp10}, the process of collecting, filtering, and annotating sign language data remains a labor-intensive and resource-demanding process. Given the requirement for proficient signers to ensure annotation accuracy, the scalability of such efforts is inherently limited. Social media focused efforts like YouTube-SL-25 \citep{tanzer2025youtubesl} have paved the way toward large-scale, open-domain multilingual SLT datasets, offering a partial solution to the challenges of scale and accessibility. Yet, the rapid development of foundation models—especially Vision-Language Models (VLMs)—presents an underexplored opportunity to further ease and scale the acquisition and annotation process by leveraging the vast amount of user-generated sign language content on platforms such as YouTube, TikTok, Pinterest, and Instagram.

Building on this direction, recent advancements in sign language translation have increasingly relied on self-supervised pre-training followed by task-specific fine-tuning. Studies such as \citep{hamidullah-etal-2024-sign, rust-etal-2024-towards, liang2024llavasltvisuallanguagetuning, zhang2024scaling, li2025unisign} have demonstrated the effectiveness of this approach, enabling models to learn meaningful representations from unlabeled data before being adapted to downstream tasks. Furthermore, even pretraining on weakly labeled data has been shown to enhance model performance, highlighting the potential of leveraging large-scale but noisier supervision signals \citep{rust-etal-2024-towards}. 

To this end, we introduce a novel vision-language model-assisted pipeline to streamline data acquisition and filtering for SLT, with a particular focus on social media platforms such as TikTok. As illustrated in Figure~\ref{fig:pipeline}, the core idea behind our framework is straightforward: leveraging the distinction between manual (hand movements) and non-manual (facial expressions) components of sign language~\cite{NUNEZMARCOS2023118993}, and observing similar structural patterns in social media content (e.g., the presence of video descriptions, hashtags, comments, and sometimes captions), we design a three-stage pipeline. Our pipeline uses the \texttt{Qwen2.5-VL} model to process videos in three stages: (1) it checks whether the subject’s face is clearly visible (\textbf{FaceDetector}) and whether the person exhibits signs of signing activity (\textbf{SignActivityDetector}); (2) it extracts visible on-screen text, which often corresponds to the creator’s intended translation (\textbf{TextExtractor}); and (3) it evaluates the alignment between the extracted text and the signing using \texttt{Phi-4-Multimodal}~\cite{microsoft2025phi4minitechnicalreportcompact} as a model-as-a-judge (\textbf{Judge}). In all, our contributions are as follows:
\begin{itemize}[itemsep=2pt, topsep=10pt, leftmargin=*]
    \item We present the first VLM-assisted pipeline for SLT data curation from social media, with a focus on TikTok. Our method focuses on capturing the core elements of sign language: face visibility, signing activity, and the corresponding spoken language translation.
    \item We release the TikTok-SL-8 dataset, comprising approximately 49 hours of video across 8 sign languages, curated through an automated VLM-assisted pipeline. Our method achieves an accuracy of 0.75 on DGS and 0.82 on ASL, closely matching human annotator performance.
    \item We additionally evaluate our dataset on DGS and ASL using two off-the-shelf SLT models to assess its performance.
\end{itemize}

\section{Related Work}
In this section, we review existing SLT datasets and methodologies for translating sign languages into spoken language text. Additionally, we highlight the emerging role of LLM-assisted techniques in data curation and filtering.

\subsection{Sign Language Translation Datasets}

One of the most widely used sign language translation datasets is RWTH-PHOENIX-2014T \citep{Camgoz_2018_CVPR}, which contains 11 hours of German Sign Language (DGS) weather forecasts interpreted from the German TV station PHOENIX. In recent years, more TV broadcast datasets have emerged, expanding the scope of sign language research. SWISSTXT \citep{camgoz2021content4allopenresearchsign} provides 152 hours of news and weather programs interpreted into Swiss German Sign Language, while the BOBSL \citep{Albanie2021bobsl} dataset contains 1,447 hours of British Sign Language (BSL) interpretations from various BBC programs across multiple domains. These datasets play a crucial role in advancing sign language translation by providing large-scale, real-world training resources. Other datasets involve signers translating predefined phrases in controlled environments or using personal recording devices. CSL-Daily~\citep{Zhou2021ImprovingSL}, comprising 23 hours of content, focuses on daily life phrases in Chinese Sign Language (CSL), while How2Sign~\citep{Duarte_CVPR2021}, with over 80 hours of data, provides instructional monologues in ASL. Moreover, multilingual sign language datasets are emerging, though many remain limited in scope and domain. AfriSign \citep{gueuwou2023afrisign} and JWSign \citep{gueuwou-etal-2023-jwsign} focus on Bible translations, while SP-10 \citep{yin_sp10} includes 10 sign languages but primarily features very short sentences. In contrast, YouTube-SL-25 \citep{tanzer2025youtubesl} is a large-scale, open-domain dataset with over 3,000 hours of content across more than 25 sign languages.

\subsection{Sign Language Translation Models}




Traditional SLT models have relied on gloss annotations as an intermediate representation, but recent advances have explored end-to-end models, self-supervised learning, and the integration of large language models (LLMs). 

\citet{Camgoz_2018_CVPR} introduced the task of SLT as an end-to-end solution that jointly incorporated glosses and evaluated their method on the PHOENIX14T dataset. They later extended it to a gloss-free model \citep{Camgoz_2020_CVPR} that directly mapped sign videos to spoken sentences. As gloss annotation is labor-intensive, research has shifted towards gloss-free SLT. \citet{slt-how2sign-wicv2023} furthered this by using I3D visual features \citep{Carreira2017QuoVA} to train an SLT model on the How2Sign dataset. 
Additionally, some works have framed SLT as a self-supervised learning problem, including sentence-level embedding-based supervision \citep{hamidullah-etal-2024-sign}, scaling datasets and models \citep{zhang2024scaling}, and addressing data scarcity concerns in large-scale privacy-aware SLT training \citep{rust-etal-2024-towards}. \citet{Radford2021LearningTV} leveraged visual-language pretraining by integrating masked self-supervised learning with CLIP to enable gloss-free SLT. Recent SLT research has increasingly incorporated advancements in LLMs and multimodal learning. The SignLLM framework \citep{Gong2024LLMsAG} transforms sign videos into a structured representation for improved processing by LLMs. LLaVA-SLT \citep{liang2024llavasltvisuallanguagetuning} employs large multimodal models by first performing linguistic continued pre-training on a sign language corpus to refine linguistic capabilities. This is followed by contrastive learning to align visual and textual representations, and finally, integration via a lightweight multi-layer perceptron connector for SLT. While prior work has focused on using LLMs for translation, to the best of our knowledge, we are the first to leverage LLM-assisted methods for data collection and annotation in the SLT domain.

\subsection{LLM-assisted Data Collection}


Collecting and annotating large-scale multilingual sign language datasets remains a significant challenge due to the complexity, subjectivity, and resource-intensive nature of the process. Traditional methods rely heavily on human annotators with domain expertise, making large-scale data curation costly and time-consuming. Recent advancements in multimodal large language models (MLLMs), such as Qwen-2.5-VL \citep{Qwen2.5-VL}, Phi-4-multimodal \citep{microsoft2025phi4minitechnicalreportcompact} and DeepSeek-VL2 \citep{wu2024deepseekvl2mixtureofexpertsvisionlanguagemodels}, present a novel opportunity to automate this process by either supplementing or fully replacing human annotators with the goal of minimizing reliance on manual labor. While such approaches have been successfully explored in other domains, they remain untapped in the context of SLT data collection and curation. While not situated in the SLT domain, several recent works have explored the use of LLMs to assist with multimodal data collection and curation. For example, \citet{choi2024voldogerllmassisteddatasetsdomain} introduces the \textsc{VolDoGer} dataset, which leverages multimodal LLM-based annotation to streamline data labeling for vision-language tasks like image captioning, visual question answering (VQA), and visual entailment (VE), reducing reliance on human annotators. \citet{wang2024modelintheloopmiloacceleratingmultimodal} introduces the Model-in-the-Loop (MILO) framework, which integrates LLMs into the data annotation process to improve efficiency and quality by assisting human annotators. They experiment with a multimodal dataset of social media comments and demonstrated that MILO reduced annotation time and enhanced labeling efficiency. \citet{yeh-etal-2024-cocolofa} collected the \textsc{CoCoLoFa} dataset with the assistance of crowd workers and LLMs.




\begin{figure}[!htbp]
\begin{center}
   \includegraphics[width=\linewidth]{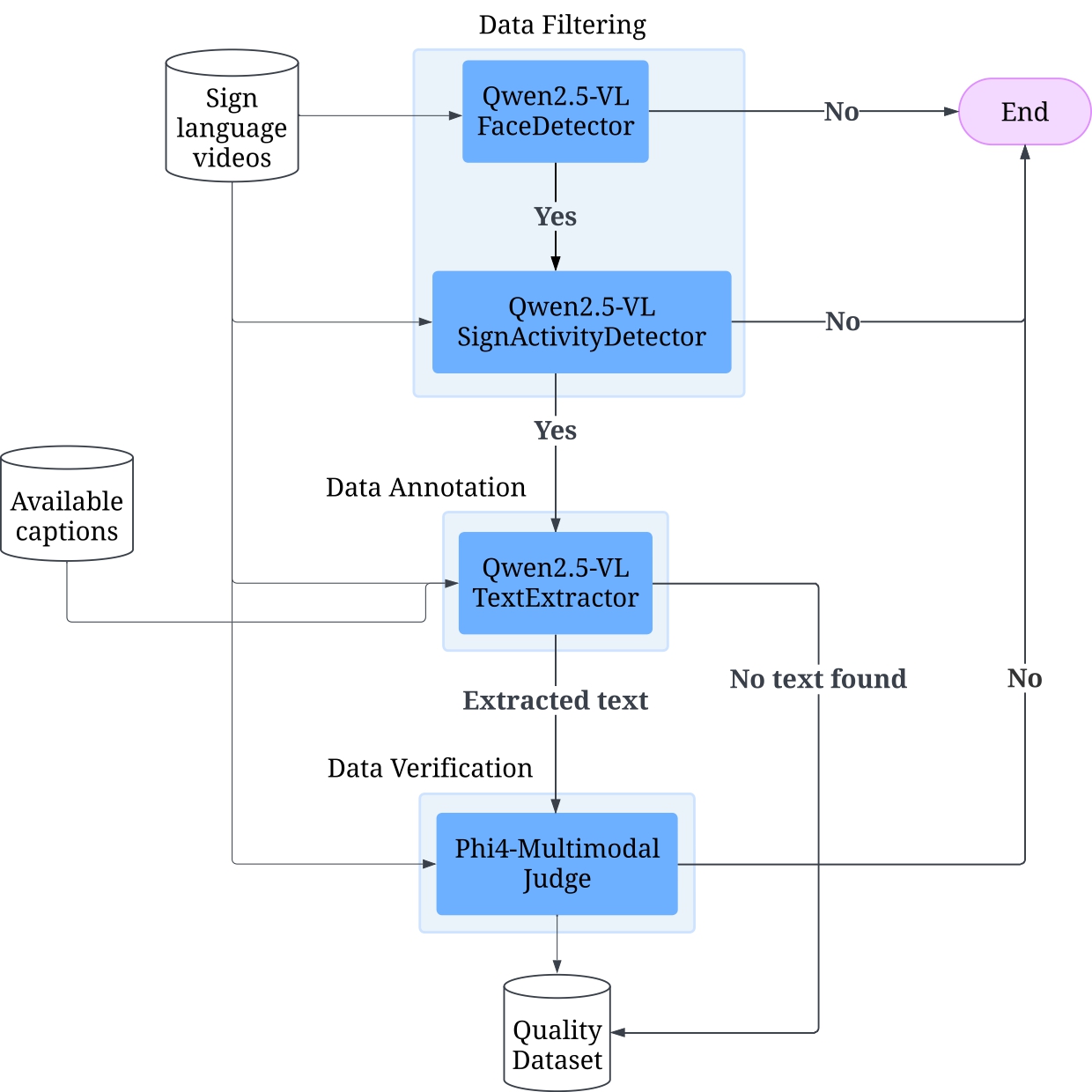}
\end{center}
\caption{An overview of our VLM-based SLT dataset collection framework on social media, with a particular focus on TikTok. The pipeline consists of three key stages: data filtering, data annotation, and data verification.}
\label{fig:pipeline}
\end{figure}

\section{Method}    

In this section, we introduce our framework, which leverages the multimodal large language model \texttt{Qwen-2.5-VL} for \textbf{filtering} and \textbf{annotation} and \texttt{Phi-4-multimodal} as a \textbf{model-as-a-judge}, as illustrated in Figure \ref{fig:pipeline}. After initial data collection, our framework consists of three key stages: Data Filtering, Annotation, and Verification. In the following sections, we will explore each stage in detail.

\subsection{Data Collection}
\label{sec:data_collection}

As motivated in the introduction, our approach adopts an social media data source setting similar to YouTube-SL-25, but extends it to a more challenging platform—TikTok—which contains potentially noisier content. Unlike YouTube, which primarily features longer, educational content, TikTok is characterized by short-form, trend-driven videos. This shift introduces additional challenges, including inconsistent caption quality and greater variability in content styles. 

\noindent Our goal is to develop a generalizable framework for SLT data acquisition and filtering. Based on our observation that social media videos tend to follow similar structural patterns---such as the presence of video descriptions, hashtags, comments, and occasionally captions---we implement two starting point strategies for automatic data collection:
\begin{enumerate}
    \item \textbf{Hashtag-based retrieval},\footnote{We provide the full list of hashtags in Appendix \ref{sec:tiktok_hashtags}} where we identify relevant hashtags.
    \item \textbf{User-based retrieval}, where we identify high-quality users who consistently produce sign language content.
\end{enumerate}

\noindent We manually select hashtags by inspecting TikTok video descriptions, ensuring that each hashtag appears in both English and the native language associated with the target sign language (e.g., for DGS, hashtags include both "Germansignlanguage" and "Gebärdensprache"). High-quality users are identified through manual inspection of profiles in TikTok’s top recommendations. Due to privacy concerns, we do not disclose the usernames of these individuals.


\subsection{Data Filtering}
\begin{figure}[!h]
\begin{center}
   \includegraphics[width=0.99\linewidth]{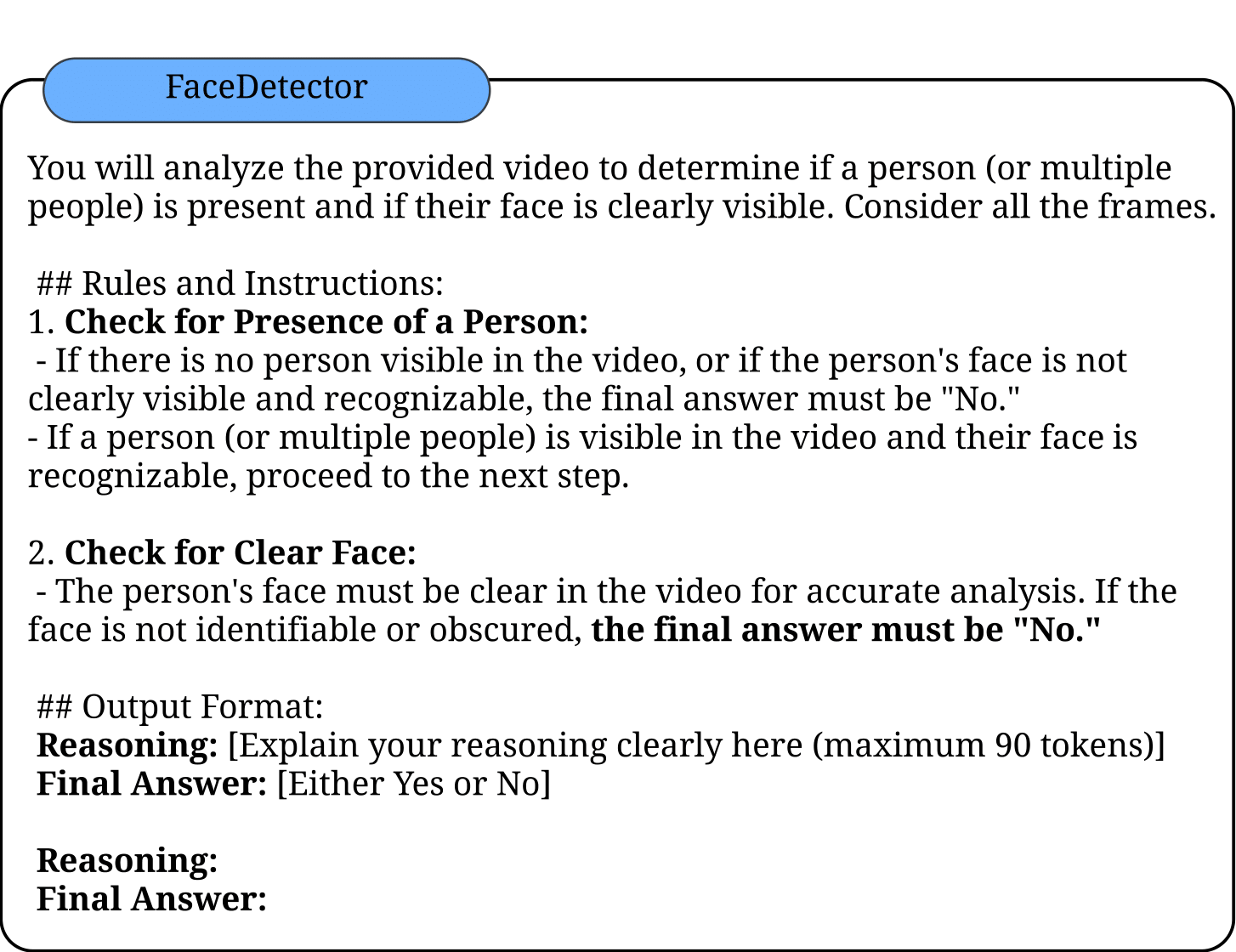}
\end{center}
\caption{Prompt template used for the \textbf{VLM FaceDetector}. }
\label{fig:face_detector_prompt}
\end{figure}

We design and rely on well-crafted prompts for VLM-based filtering in order to automate and scale the identification of sign language unrelated content. To this end, we use \texttt{Qwen-2.5-VL} as the VLM-based \textbf{FaceDetector} and \textbf{SignActivityDetector}, due to its strong performance on video understanding tasks \citep{Qwen2.5-VL}. The VLM \textbf{FaceDetector} analyzes the video frames to determine if a person (or multiple people) is present and whether their face is clearly visible and identifiable. If the face is obscured or unrecognizable, the video will be discarded. The specific prompt we used for the FaceDetector stage is shown in Figure \ref{fig:face_detector_prompt}. The VLM \textbf{SignActivityDetector}, on the other hand, assesses whether the person in the video is primarily using sign language for communication, classifying individuals based on their signing behavior. \footnote{We provide the prompt used for SignActivityDetector stage in Appendix \ref{app:prompt_template}.} Together, these two stages help filter videos to ensure relevant and high-quality sign language content is selected.

\subsection{Data Annotation}

Unlike other datasets discussed in the related work section—such as those sourced from YouTube, which often come with seemingly well-aligned captions, or those collected in controlled lab environments with invited signers, resulting in high-quality but small and less diverse datasets—TikTok presents a unique set of challenges for sign language translation data curation. TikTok videos typically lack formal captioning. Instead, creators often embed text directly within the video itself. However, this text is not always a faithful translation of the signed content; it may simply include the title of a lesson, a promotional link, or the name of a song being signed. This ambiguity complicates the task of aligning signed content with corresponding text. To address these challenges while benefiting from TikTok’s diversity and real-world context, we again use \texttt{Qwen2.5-VL} for its video optical character recognition (OCR) and understanding capabilities. Specifically, we use a vision-language module called \textbf{TextExtractor}, which identifies and extracts text from videos featuring sign language. It verifies the extracted text for readability and alignment with the video content, prefers formal captions when available, and flags instances where no textual content is found. \footnote{The prompt template used for the VLM TextExtractor is available in Appendix \ref{app:prompt_template}.}


\subsection{Data Verification}

To ensure the accuracy and relevance of our framework, we adopt a \say{model-as-a-judge} approach using a VLM \textbf{Judge} as the final verification step. As previously discussed, while some TikTok videos include captions, these captions may not accurately reflect the signed content. In many cases, the text embedded within the video may be unrelated—serving instead as a lesson title, app promotion, or song title—rather than a true translation of the signs. To address this issue, we employ \texttt{Phi-4-Multimodal} as a VLM \textbf{Judge}. In accordance with best practices in \say{model-as-a-judge} approaches, and to mitigate common problems such as \textit{self-preference bias} \citep{wataoka2024selfpreferencebiasllmasajudge} and \textit{preference leakage} \citep{li2025preferenceleakagecontaminationproblem}, we intentionally use a model different from \texttt{Qwen2.5-VL} for the evaluation stage. \texttt{Phi-4-multimodal} has also demonstrated stronger performance than \texttt{Qwen2.5-VL} on both the multi-image benchmark \textsc{BLINK}~\cite{fu2024blink} and the video benchmark \textsc{VideoMME}~\cite{fu2024video}, making it well-suited to analyze the signed content, determine its meaning, and verify whether it matches the provided caption—ensuring the resulting dataset remains both accurate and reliable. \footnote{The prompt template employed for the VLM Judge is available in Appendix \ref{app:prompt_template}.}

\subsection{Resulting Dataset}

As highlighted in the introduction, in terms of content we adopt an social media setting similar to YouTube-SL-25. Our goal in building \textbf{TikTok-SL-8} is to reduce the manual effort typically required from human annotators by leveraging the capabilities of multimodal LLMs. The final dataset, constructed using our fully automatic three-stage framework across eight sign languages, is summarized in Table~\ref{tab:sign-language-stats}, which presents the total number of videos and hours of content per sign language. In total, TikTok-SL-8 comprises \textbf{49 hours} of sign language video content across 8 languages sourced from TikTok---surpassing the SP-10 dataset, which contains 14 hours across ten sign languages. Notably, the entire dataset was curated automatically using our pipeline. The only manual interventions involved were in selecting relevant hashtags, identifying high-quality content creators (see Section \ref{sec:data_collection}), and crafting prompt templates tailored to each stage of the framework.

\begin{table}[h!]
\centering
\resizebox{0.99\linewidth}{!}{%
\begin{tabular}{lccc}
\toprule
\textbf{Sign Language} & \textbf{ISO 639} & \textbf{\#Videos} & \textbf{\#Hours} \\
\midrule
American    & \texttt{ase} & 816  & 8  \\
Australian  & \texttt{asf} & 541  & 4  \\
British     & \texttt{bfi} & 1201 & 15 \\
Chinese     & \texttt{csl} & 218  & 2  \\
French      & \texttt{fsl} & 285  & 3  \\
German      & \texttt{gsg} & 499  & 4  \\
Italian     & \texttt{ise} & 678  & 7  \\
Swedish     & \texttt{swl} & 562  & 6  \\
\midrule
Total & --- & 4800 & 49 \\
\bottomrule
\end{tabular}%
}
\caption{Statistics for different sign languages in the curated dataset, including ISO 639 codes, total number of videos, and total duration.}
\label{tab:sign-language-stats}
\end{table}

\section{Experiments}

We demonstrate the value of our VLM-based data acquisition and curation framework, as well as the quality of the resulting dataset for SLT, across two sign languages. Additionally, we describe how our automatic data curation framework was applied to YouTube-SL-25 (DGS) and evaluate its performance using a manually curated subset of the data as gold-standard annotations.

\subsection{Implementation Details}
For \texttt{Qwen2.5-VL}, we use the 7B variant, and for \texttt{Phi-4-multimodal}, we employ the \texttt{Phi-4-multimodal-instruct} model with 5.57B parameters available on Hugging Face. To process the videos, we adopt the default smart resizing method\footnote{\url{https://github.com/QwenLM/Qwen2.5-VL/blob/main/qwen-vl-utils/src/qwen_vl_utils/vision_process.py}} provided by the \texttt{Qwen2.5-VL} series, which supports dynamic resolution and frame rate. In our setup, however, we fix the input resolution to a width and height of 224 pixels.

\subsection{Data Acquisition and Curation Evaluation}

To validate the quality of our framework, the first author\footnote{The first author is a non-native hearing annotator. Annotations were performed by comparing the signs in the video with any captions provided within the video, the automatic captions (if available), and user comments to assess consistency and semantic alignment.} manually annotated approximately 150 samples from the DGS and ASL TikTok-SL-8 subsets included in the final dataset. We use Accuracy, Precision, and Recall to evaluate the overall performance of our framework in comparison to manual annotation. Furthermore, to demonstrate the applicability of our approach to other open-domain settings such as YouTube, we further evaluate the framework by applying it to the already curated YouTube-SL-25 (DGS). Table~\ref{tab:framework-performance} presents the performance of our VLM-based framework across different datasets and sign languages. For the YouTube-SL-25 (DGS) gold dataset, we treat all samples as positive ground-truth labels, which makes precision not applicable. On the gold DGS data set, our framework achieves an accuracy and recall of 0.86. On our TikTok-SL-8 dataset, the framework shows solid performance across both DGS and ASL, with particularly high precision on ASL (0.91) and strong recall on DGS (0.87), indicating its effectiveness in diverse and noisy open-domain scenarios.

\begin{table}[h]
\centering
\resizebox{0.98\linewidth}{!}{%
\begin{tabular}{lccc}
\hline
\addlinespace[0.7ex]  
\textbf{Dataset} & \textbf{Accuracy} & \textbf{Precision} & \textbf{Recall} \\
\addlinespace[0.7ex]  
\hline
\addlinespace[0.7ex]  
YouTube (DGS) & 0.86 & --   & 0.86 \\
\addlinespace[0.4ex]  
TikTok (DGS)        & 0.75 & 0.72 & 0.87 \\
\addlinespace[0.4ex]  
TikTok (ASL)        & 0.82 & 0.91 & 0.79 \\
\addlinespace[0.7ex]  
\hline
\end{tabular}}
\caption{Performance of our framework across TikTok-SL-8 (TikTok) and YouTube-SL-25 (YouTube). Precision is not applicable to the YouTube dataset due to the absence of negative ground-truth labels.}
\label{tab:framework-performance}
\end{table}

\begin{figure}[!htbp]
\begin{center}
   \includegraphics[width=\linewidth]{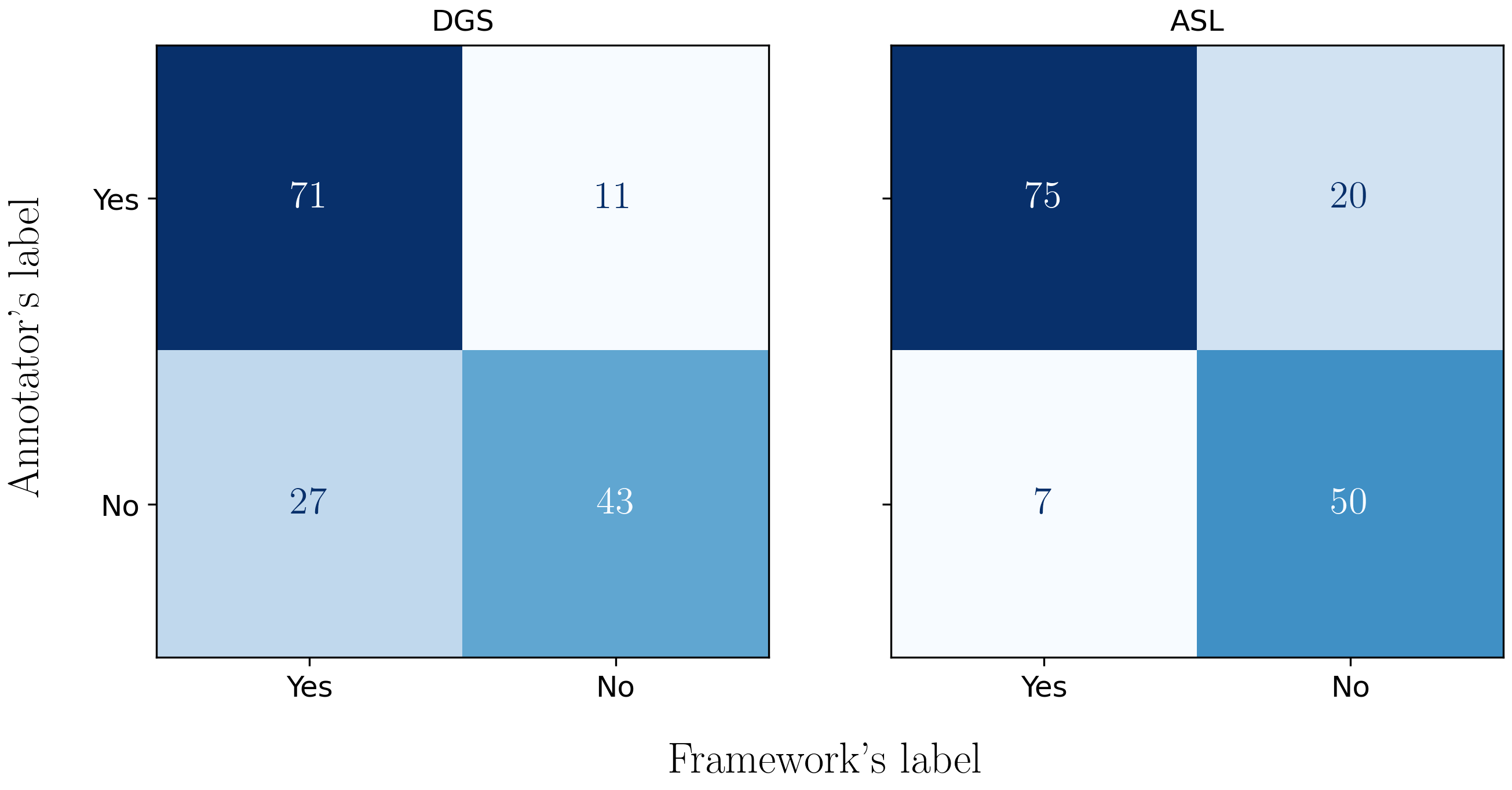}
\end{center}
\caption{Confusion matrices showing the agreement between our framework's automated labels and manual annotations for DGS and ASL subsets of TikTok-SL-8.}

\label{fig:confusion_matrix}
\end{figure}

\noindent Figure~\ref{fig:confusion_matrix} presents the confusion matrices between our framework’s predicted labels and manual annotations on the DGS and ASL subsets of the TikTok-SL-8 dataset. The confusion matrices show strong agreement, particularly for ASL, where the framework achieves 75 true positives and 50 true negatives, indicating high precision and recall. In DGS, the framework also performs well, with 71 true positives, though with a slightly higher false positive count.

\subsection{SLT Experiments Using the Data}
\label{sec:experimental_settings}

\textbf{Baselines.} We demonstrate the effectiveness of our VLM-based data acquisition and curation framework and the TikTok-SL-8 dataset using two gloss-free open-source baselines across two sign languages: SEM-SLT \citep{hamidullah-etal-2024-sign} and Signformer \citep{eta2024signformer}. 

\paragraph{SEM-SLT \citep{hamidullah-etal-2024-sign}.}

SEM-SLT consists of a visual encoder that maps videos to sentence embeddings (sign2sem) and a decoder that maps sentence embeddings to text (sem2text). Both components are pretrained separately and then combined and fine-tuned in an end-to-end fashion. We train the sign2em visual encoder to map sign language videos into  sentence embeddings using video features extracted with EF-Net-B0 \cite{pmlr-v97-tan19a} from the TikTok-SL-8 dataset. The sentence embeddings are intended to represent the overall meaning of the signed utterances. \footnote{More details about the SEM-SLT architecture can be found in Appendix~\ref{appendix:sem-slt}.}

\noindent The original sem2text module, which maps semantic embeddings to spoken language sentences, was designed for single-sentence outputs. However, since TikTok-SL-8 contains multi-sentence segments, we adapted the sem2text decoder accordingly. We first pre-trained the embedding projection layer and the first six layers of the mBART \cite{liu-etal-2020-multilingual-denoising} decoder on the CNN/Daily Mail and MLSUM[de] datasets \citep{see-etal-2017-get, scialom2020}. This pretraining step allowed the model to better handle longer text sequences. We then fine-tuned the mBART decoder on the target transcriptions from TikTok-SL-8 training data, aligning it with the TikTok-specific language and structure. Finally, we trained the full SEM-SLT pipeline in an end-to-end manner. This involved combining the visual encoder and the sem2text decoder into a single model—Sign2(sem+text)—and supervising it directly using the target TikTok-SL-8 training data sentence embeddings. This step refines the entire system to translate raw sign language video into spoken language text.

\paragraph{Signformer \citep{eta2024signformer}.} 
Signformer is an encoder--decoder Transformer-based model designed for edge AI applications, achieving strong performance with a compact size of only 0.57 million parameters. Like \citet{hamidullah-etal-2024-sign} SEM-SLT, Signformer is a gloss-free model that does not rely on any pre-training. Instead, it leverages a novel attention mechanism proposed by \citet{10205495}, combined with a convolutional module. We train Signformer from scratch using features extracted by the S3D model \citep{Xie_2018_ECCV}, pre-trained on both the WLASL \citep{li2020word} and Kinetics-400 \citep{kay2017kineticshumanactionvideo} datasets. For feature extraction, we use only the first four blocks of the S3D architecture. Each input video is passed through the S3D encoder, and the output from the final block is spatially pooled to obtain a feature representation of size \( F/4 \times 832 \), where \( F \) denotes the number of frames in the video.

\paragraph{Evaluation Metrics.} We evaluate translation performance using standard metrics: chrF \citep{popovic-2015-chrf}, BLEU \citep{papineni-etal-2002-bleu}, and BLEURT \citep{sellam-etal-2020-bleurt}. For BLEU\footnote{\texttt{BLEU|nrefs:1|case:mixed|eff:no|tok:13a| smooth:exp|version:2.5.1}} and chrF,\footnote{\texttt{chrF|nrefs:1|case:mixed|eff:yes|nc:6|nw:0| space:no|version:2.5.1}} we use sacreBLEU \citep{post-2018-call}. For BLEURT, we use the official Python library.\footnote{BLEURT using checkpoint BLEURT-20.}

\subsection{Results on TikTok-SL-8}

To demonstrate the quality of our VLM-based data collection and curation framework, we evaluate on held out test sets from the ASL and DGS subsets of our resulting dataset, TikTok-SL-8. We consider only those videos that either include captions or for which our framework successfully extracted text. After this filtering step, the ASL subset contains 649 training and 75 test videos, while the DGS subset contains 391 training and 71 test videos. We report results using Signformer and SEM-SLT using the BLEURT, chrF, and BLEU metrics. Table \ref{tab:metric-results} presents the quantitative results. The results show that SEM-SLT, a model incorporating pre-training, consistently outperforms Signformer, even when trained on noisy data that includes misaligned captions. While BLEURT and chrF scores are relatively close for both models, BLEU scores indicate a clear advantage for SEM-SLT. Overall, the findings show that, despite being collected through an automatic, VLM-assisted pipeline and containing noisy supervision, the TikTok-SL-8 dataset retains sufficient alignment between sign and text to support effective training of sign language translation models. Overall, results indicate that the pretraining-based SEM-SLT model is better able to handle slightly noisy data than Signformer.

\begin{table}[!htbp]
\centering
\resizebox{0.99\linewidth}{!}{%
\begin{tabular}{llccc}
\hline
\addlinespace[0.7ex]  
\textbf{Dataset} & \textbf{Method} & \textbf{BLEURT} & \textbf{chrF} & \textbf{BLEU} \\
\addlinespace[0.7ex]  
\hline
\addlinespace[0.7ex]  
    TikTok (DGS) & Signformer & $0.18 \pm 0.04 $ & $16.1 \pm 6.7 $ & $5.2 \pm 4.9$ \\
\addlinespace[0.5ex]  
             & SEM-SLT    & $0.21 \pm 0.04 $ & $18.8 \pm 5.6 $ & $9.2 \pm 5.9$ \\
\addlinespace
TikTok (ASL) & Signformer & $0.30 \pm 0.03$ & $13.1 \pm 2.5$ & $0.9 \pm 0.7$ \\
\addlinespace[0.5ex]  
             & SEM-SLT    & $0.31 \pm 0.02 $ & $ 15.8 \pm 3.6 $ & $5.6 \pm 4.3$ \\
\addlinespace[0.7ex]  
\hline
\end{tabular}%
}
\caption{Performance of Signformer and SEM-SLT on the DGS and ASL subsets of TikTok-SL-8 (TikTok) using BLEURT, chrF, and BLEU.}
\label{tab:metric-results}
\end{table}






\begin{table}[!h]
\centering
\resizebox{0.99\linewidth}{!}{%
\begin{tabular}{c|p{11cm}}
\hline
\addlinespace[0.7ex]  
\textbf{Lang.} & \textbf{Text (German / English)} \\
\addlinespace[0.7ex]  
\hline
\addlinespace[0.7ex]  
\multirow{12}{*}{DGS} 
& \textbf{Ref:} Hallo! Du hast bis jetzt nur Lautsprache gelernt. Französisch, Spanisch und so? Und jetzt noch mal eine Lautsprache lernen?. Nein, jetzt ist Gebärdensprache dran! \\
& \textit{(Hello! Up until now, you've only learned spoken languages. French, Spanish, and so on? And now you're going to learn another spoken language? No, now it's time for sign language!)} \\
& \textbf{Signformer:} ich liebe dich. \\
& \textit{(I love you.)} \\
& \textbf{SEM-SLT:} Hallo! Du hast bis jetzt nur Lautsprache gelernt. Französisch, Spanisch und so? Und jetzt noch mal eine Lautsprache lernen? Nein, jetzt ist Gebärdensprache dran! \\
& \textit{(Hello! Up until now, you've only learned spoken languages. French, Spanish, and so on? And now you're going to learn another spoken language? No, now it's time for sign language!)} \\
\addlinespace[0.7ex]  
\cline{2-2}
\addlinespace[0.7ex]  
& \textbf{Ref:} Mein Name Mein Name Alles gut. Alles gut. gut gut Dein Name was? Dein Name was? \\
& \textit{(My name, my name. All good. All good. Good, good. Your name what? Your name what?)} \\
& \textbf{Signformer:} alles gut? gut. dein name was? mein name. \\
& \textit{(All good? Good. Your name what? My name.)} \\
& \textbf{SEM-SLT:} Ich bin froh, dass du hier bist. \\
& \textit{(I am glad that you are here.)} \\
\addlinespace[0.7ex]  
\hline
\addlinespace[0.7ex]  
\multirow{6}{*}{ASL} 
& \textbf{Ref:} The sign of the day is: try, attempt, effort. Try. For example, "If, at first, you don't succeed. Try, try again."
\\
& \textbf{Signformer:} the sign of the day is ongoing or continue. on go. ongoing. for example, "you want to become proficient at american sign language? well, i'll tell you how to do it socialize with deaf individuals, pick up on the signs that they use, and accept their feedback. it's an ongoing process."
 \\
& \textbf{SEM-SLT:} ASL is more. Try. For example, maybe more. Try. For example, at the end of the day you're gonna try to try again. Try my body is shorten to the day. Try. It'
\\
\addlinespace[0.7ex]  
\cline{2-2}
\addlinespace[0.7ex]  
& \textbf{Ref:} Today I learned School, School. \\
& \textbf{Signformer:} cc today i learned school, school. \\
& \textbf{SEM-SLT:} Today I Learned school. School \\
\addlinespace[0.7ex]  
\hline
\end{tabular}
}
\caption{Example translations from TikTok-SL-8 for DGS and ASL subsets using Signformer and SEM-SLT models.}
\label{tab:example_translations}
\end{table}

\subsection{Qualitative Analysis}
Table \ref{tab:example_translations} presents two example translations each from the DGS and ASL subsets of our TikTok-SL-8 dataset, using the Signformer and SEM-SLT models. In the DGS examples, SEM-SLT slightly outperforms Signformer in translation quality, aligning with the quantitative results shown in Table~\ref{tab:metric-results}. In the ASL examples, SEM-SLT again demonstrates more accurate translations. This aligns with the BLEU score of 5.59 reported for ASL in Table \ref{tab:metric-results}. Manual analysis confirms that the pretraing-based SEM-SLT is better able to use slightly noisy data than Signformer.

\subsection{Ablation Study}
To further assess the effectiveness of our approach, we conducted an additional ablation study on both our data collection and curation framework, and the resulting dataset.

\noindent\textbf{Caption Agreement.} Although \texttt{Qwen2.5-VL-7B} demonstrated promising OCR capabilities, we further evaluated its performance on the DGS and ASL subsets of TikTok-SL-8 by measuring caption agreement between our framework (VLM TextExtractor) and human annotator translations. We used BLEURT, chrF, and BLEU as evaluation metrics. Videos for which our framework returned \say{No text found.} were excluded from the analysis. After this filtering step, the held-out test sets contained 65 samples for the ASL subset and 66 for the DGS subset. Table~\ref{tab:caption-agreement} shows a high degree of agreement across both subsets, with notably higher alignment in the ASL subset. 


\begin{table}[!h]
\centering
\resizebox{0.99\linewidth}{!}{%
\begin{tabular}{lccc}
\hline
\addlinespace[0.7ex]  
\textbf{Dataset} & \textbf{BLEURT} & \textbf{chrF} & \textbf{BLEU} \\
\addlinespace[0.7ex]  
\hline
\addlinespace[0.7ex]  
TikTok (DGS) & $0.53 \pm 0.08$ & $66.5 \pm 13.3 $ & $52.1 \pm 17.5 $ \\
\addlinespace[0.7ex]  
TikTok (ASL) & $0.71 \pm 0.06 $ & $83.6 \pm 9.3 $ & $77.0 \pm 13.6 $ \\
\addlinespace[0.7ex]  
\hline
\end{tabular}%
}
\caption{Caption agreement between framework-extracted captions and human annotator translations on the DGS and ASL subsets of TikTok-SL-8, measured using BLEURT, chrF, and BLEU.}
\label{tab:caption-agreement}
\end{table}


\section{Conclusion}

In this paper, we introduced the first automated data acquisition and curation framework based on Vision-Language Models for sign language translation from social media. Our approach simplifies dataset collection for sign language translation and supports human annotators by implementing a three-stage pipeline. Our pipeline leverages \texttt{Qwen2.5-VL} as a FaceDetector, SignActivityDetector, and TextExtractor, and employs \texttt{Phi-4-multimodal} as a \textit{Model-as-a-Judge} to assess whether the extracted text aligns with the signing in the video. We evaluated our VLM-based data acquisition and curation framework through two complementary approaches. First, we assessed the quality of automatically extracted data by comparing it against manually annotated subsets of our TikTok-SL-8 dataset for both ASL and DGS, as well as demonstrating how our framework performs on the already curated YouTube-SL-25 gold dataset for DGS. This evaluation allows us to benchmark the effectiveness of our VLM-based pipeline. Second, we evaluated the utility of the curated data by training two open-source gloss-free SLT models—\textit{Signformer} and \textit{SEM-SLT}. Results indicate that the pretraining-based SEM-SLT model is better able to handle slightly noisy data than Signformer. Furthermore, we demonstrated that the captions extracted by our framework exhibit a high degree of agreement with human-translated references. We believe our VLM-based framework can streamline the acquisition of large amounts of sign language data already available on social media platforms.

\section*{Limitations}

While our automated data acquisition and curation framework for sign language translation from social media shows promising results, there are both limitations and opportunities for improvement. In particular, we rely on VLMs for two critical stages: sign activity recognition and text extraction. Despite notable progress in OCR tasks with VLMs, significant challenges remain---particularly in handling longer videos due to context length limitations. These limitations can lead to misclassifications, such as incorrectly labeling hearing individuals as signers, or generating captions that are poorly aligned with the actual content. Additionally, our approach primarily relied on prompt engineering, which may not yield optimal results. More sophisticated prompting strategies, such as Chain-of-Thought prompting \citep{10.5555/3600270.3602070}, could potentially enhance our framework.

\section*{Acknowledgements}
This work was partially funded by the German ministry for education and research (BMBF) through projects BIGEKO (grant number 16SV9093) and TRAILS (grant number 01IW24005).
CEB acknowledges her AI4S fellowship within the “Generación D” initiative by Red.es, Ministerio para la Transformación Digital y de la Función Pública, for talent attraction (C005/24-ED CV1), funded by NextGenerationEU through PRTR.

\bibliographystyle{acl_natbib}
\bibliography{ranlp2023}

\clearpage
\appendix

\section{Appendix}
\label{sec:appendix}

\subsection{Crawling Hashtags}
\label{sec:tiktok_hashtags}

\begin{figure}[!h]
\begin{center}
   \includegraphics[width=0.99\linewidth]{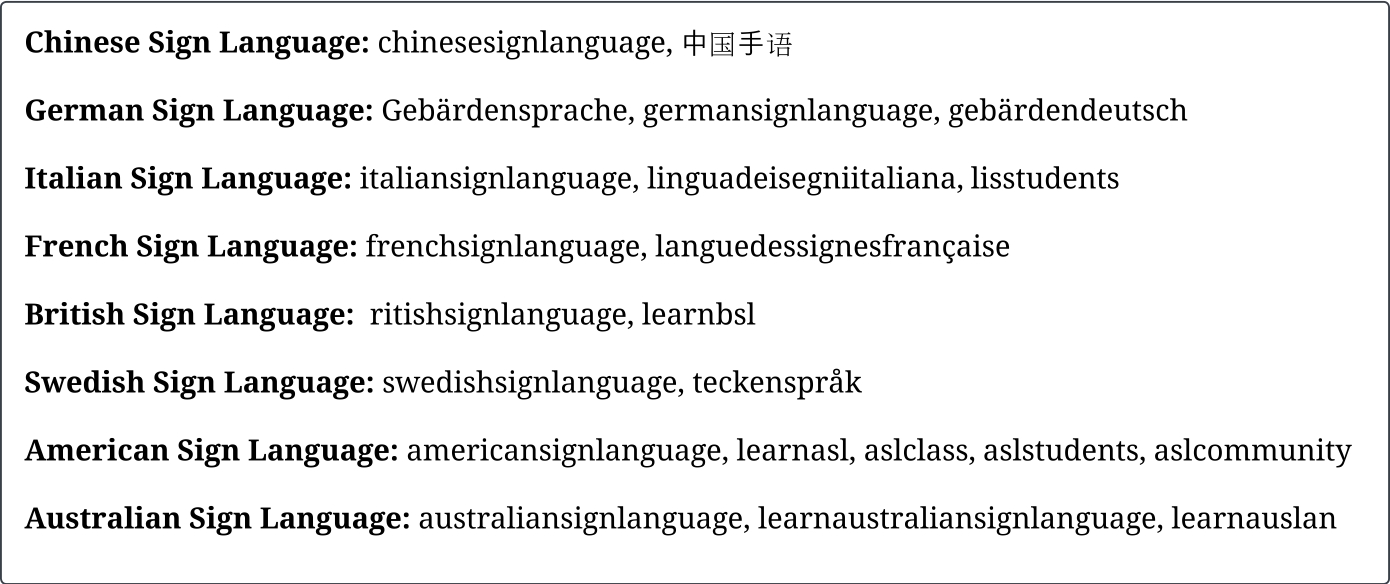}
\end{center}
\caption{Hashtags used for crawling videos for each sign language in the TikTok-SL-8 dataset.}
\label{fig:tiktok_hashtags}
\end{figure}

To collect sign language videos from TikTok, we curated a list of relevant hashtags for each of the eight sign languages included in our TikTok-SL-8 dataset. These hashtags were selected based on common usage within the TikTok and Instagram communities, along with manual validation. Figure \ref{fig:tiktok_hashtags} shows the specific hashtags used to crawl videos for each language.

\subsection{Prompt Template}
\label{app:prompt_template}

\begin{figure}[!htbp]
\begin{center}
   \includegraphics[width=0.99\linewidth]{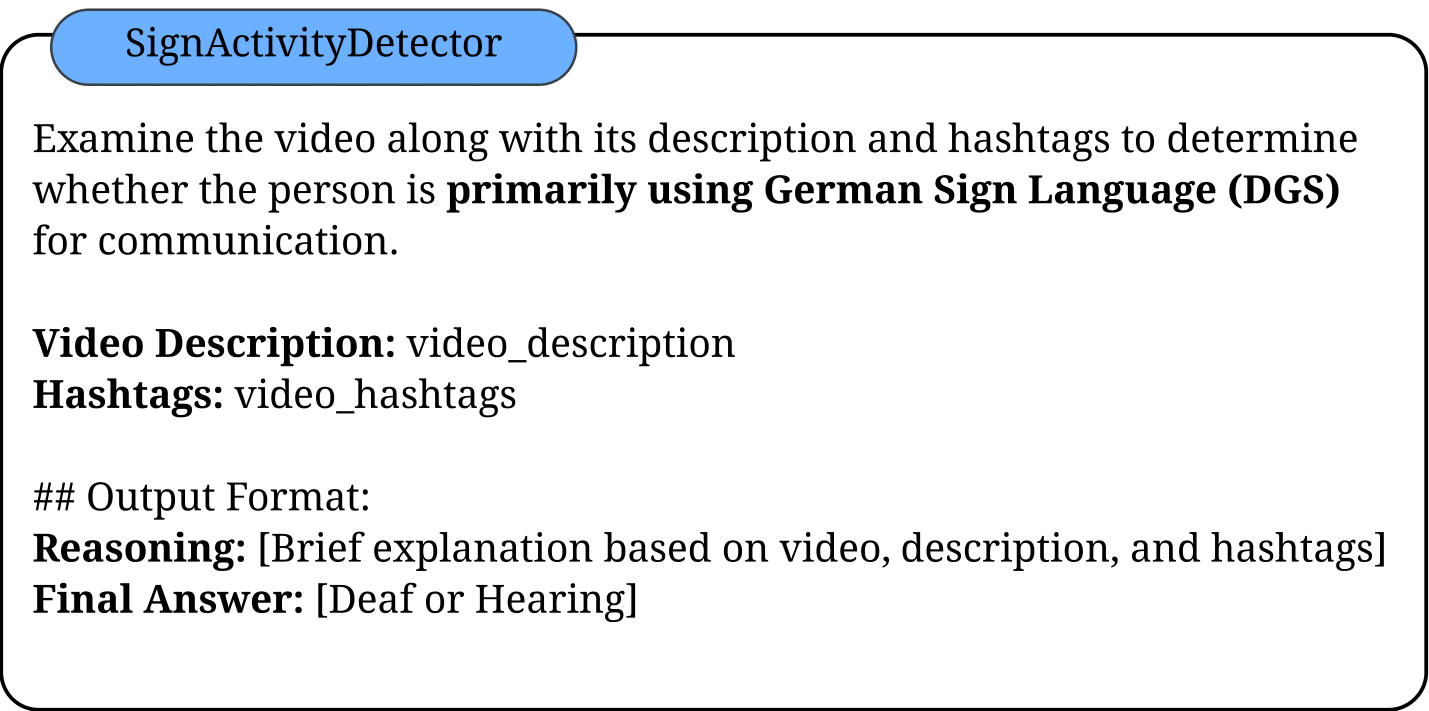}
\end{center}
\caption{Prompt template used for the VLM SignActivityDetector.}
\label{fig:signactivitydetector_prompt}
\end{figure}

\begin{figure}[!htbp]
\begin{center}
   \includegraphics[width=0.99\linewidth]{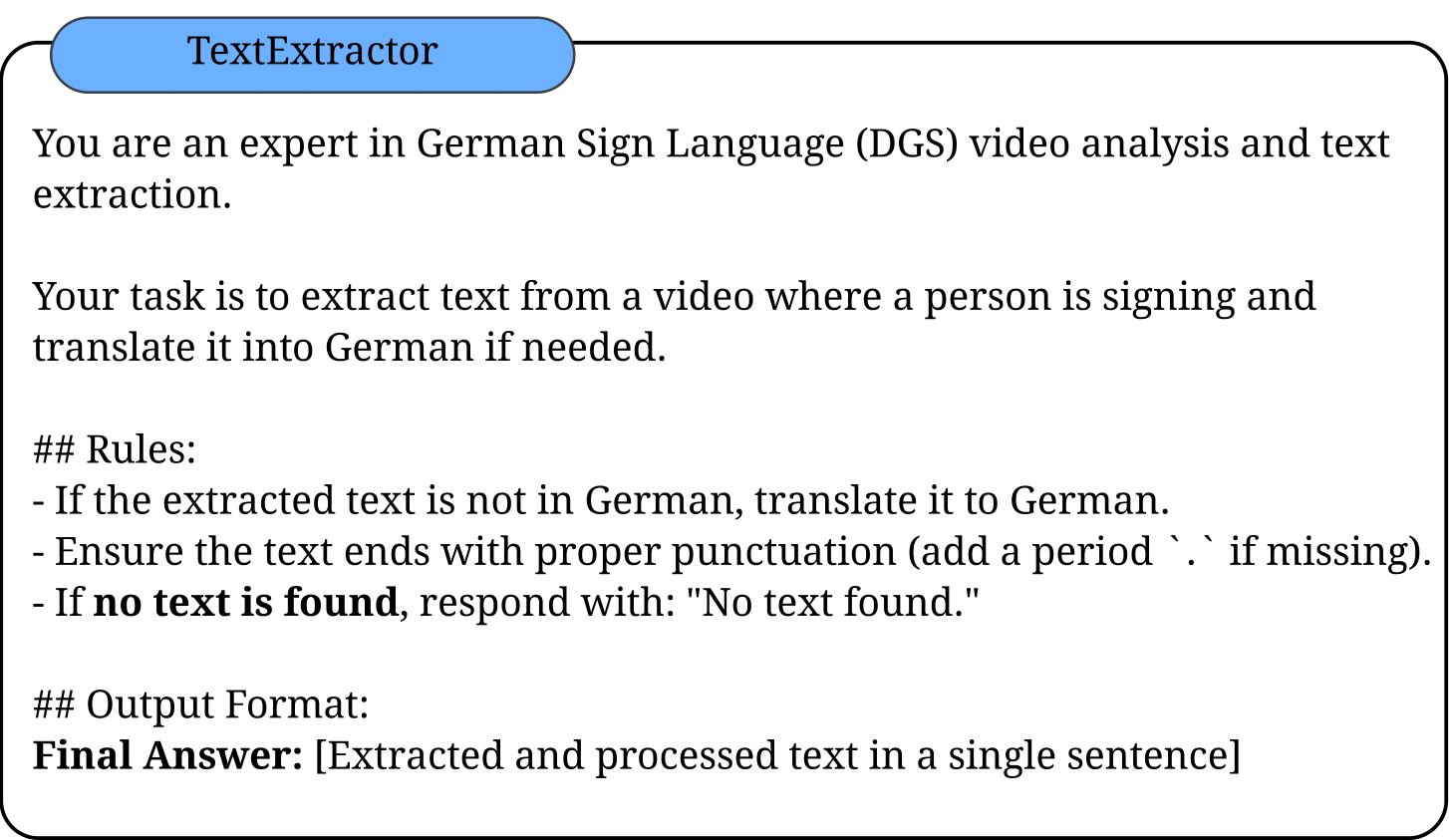}
\end{center}
\caption{Prompt template used for the VLM TextExtractor.}
\label{fig:textextractor_prompt}
\end{figure}

\begin{figure}[!htbp]
\begin{center}
   \includegraphics[width=0.99\linewidth]{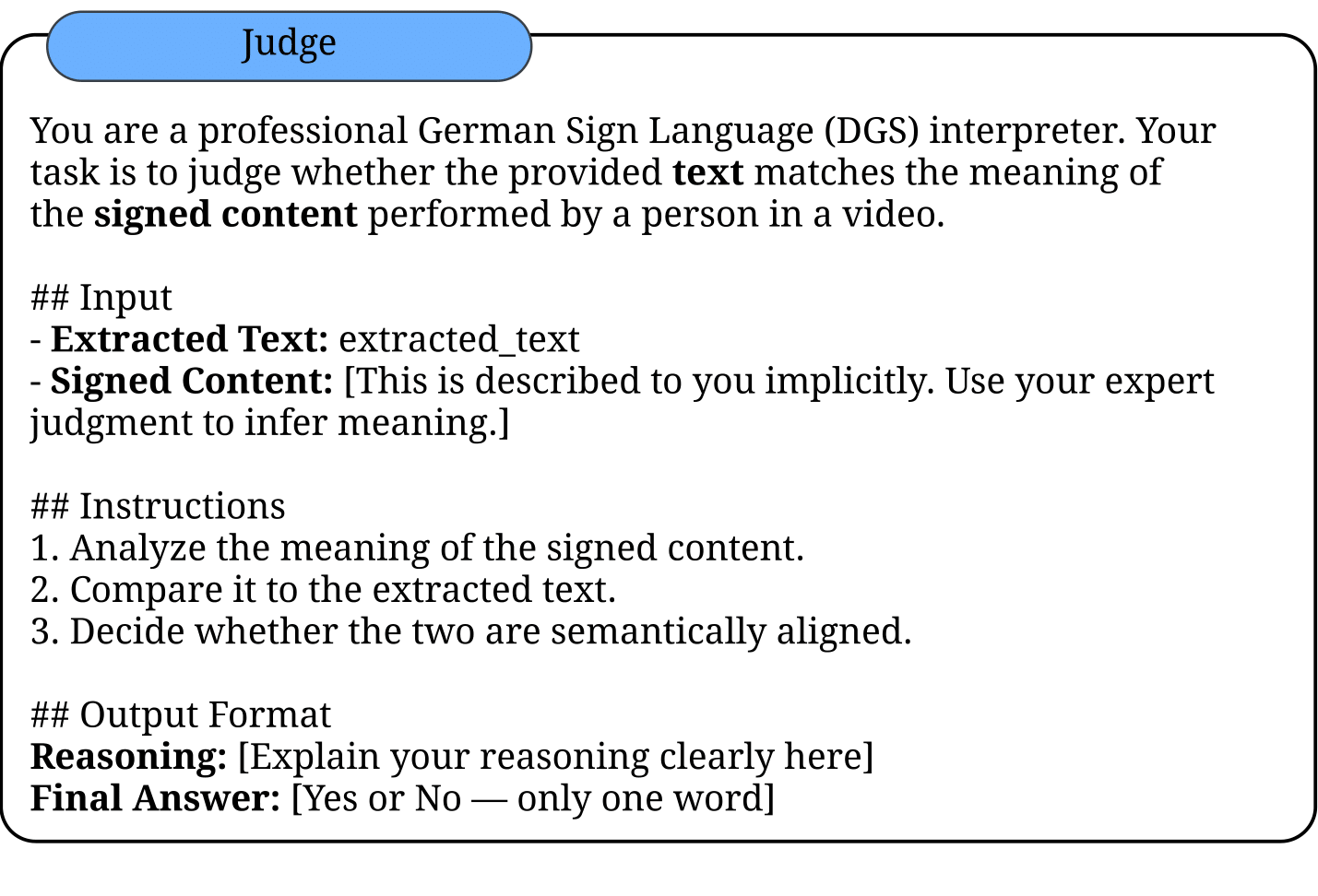}
\end{center}
\caption{Prompt template used for the VLM Judge.}
\label{fig:judge_prompt}
\end{figure}

We employ four specialized prompts throughout our framework, each designed with a clear and instructive pattern tailored to a specific task. The prompt templates used in the VLM components—SignActivityDetector, TextExtractor, and Judge—for German Sign Language (DGS) are shown in Figure \ref{fig:signactivitydetector_prompt}, Figure \ref{fig:textextractor_prompt}, and Figure \ref{fig:judge_prompt}, respectively. These prompt templates can be easily adapted to other sign languages as well.

\subsection{SEM-SLT}
\label{appendix:sem-slt}

SEM-SLT is a gloss-free sign language translation model composed of two sequentially trained modules: a sign-to-sentence embeddings encoder (\textit{sign2sem}) and a sentence embeddings-to-text decoder (\textit{sem2text}). The \textit{sign2sem} module encodes frame-level video features—extracted using \textit{EF-Net-B0}—into a fixed vector representation that captures sentence-level meaning without relying on gloss supervision. Independently, the \textit{sem2text} module is trained to generate spoken language sentences from these semantic vectors using \textit{mbart-cc-25} \citep{liu-etal-2020-multilingual-denoising}. Finally, a full end-to-end pipeline is trained by combining both modules and applying additional supervision using the target sentence embeddings to align the latent semantic space, enabling the model to effectively bridge the gap between visual input and text without intermediate gloss labels.

\end{document}